\documentclass[10pt,twocolumn,letterpaper]{article}

\usepackage{cvpr}              
\usepackage{algorithm}
\usepackage{algorithmic}
\usepackage{multirow}
\usepackage{tabularx}
\usepackage{amsfonts,bm}
\usepackage{threeparttable}
\usepackage{upgreek}
\usepackage{pifont}%
\usepackage{bbding}
\usepackage[normalem]{ulem}
\usepackage{makecell}
\usepackage{color}
\usepackage{xcolor}
\usepackage{colortbl}
\usepackage{wrapfig}
\usepackage{soul}
\usepackage{wasysym}
\usepackage{xspace}
\usepackage{graphicx,cuted}   
\usepackage{lipsum}
\usepackage{subcaption}
\usepackage[first=0,last=9]{lcg}
\usepackage{fontawesome5, hyperref}

\definecolor{Gray}{gray}{0.85}

\newcolumntype{x}[1]{>{\centering\arraybackslash}p{#1pt}}
\newcolumntype{y}[1]{>{\raggedright\arraybackslash}p{#1pt}}
\newcolumntype{z}[1]{>{\raggedleft\arraybackslash}p{#1pt}}
\definecolor{cvprblue}{rgb}{0.21,0.49,0.74}
\hypersetup{
    pagebackref,
    breaklinks,
    colorlinks,
    allcolors=cvprblue
}

\title{YOLO-Master: MOE-Accelerated with Specialized Transformers for Enhanced Real-time Detection}

\author{
    Xu Lin$^{1}$\thanks{Equal contribution.}, \quad Jinlong Peng$^{1}$\footnotemark[1],\quad Zhenye Gan$^{1}$, \quad Jiawen Zhu$^{2}$, \quad Jun Liu$^{1}$ \\
    \noalign{\smallskip}
    $^{1}$Tencent Youtu Lab \qquad $^{2}$Singapore Management University \\
    {\tt\small \{gatilin, jeromepeng, wingzygan, juliusliu\}@tencent.com} \\
    {\tt\small jwzhu.2022@phdcs.smu.edu.sg}
}

\begin{document}
\maketitle
\begin{abstract}

Existing Real-Time Object Detection (RTOD) methods commonly adopt YOLO-like architectures for their favorable trade-off between accuracy and speed. However, these models rely on static dense computation that applies uniform processing to all inputs, misallocating representational capacity and computational resources such as over-allocating on trivial scenes while under-serving complex ones. This mismatch results in both computational redundancy and suboptimal detection performance.
To overcome this limitation, we propose YOLO-Master, a novel YOLO-like framework that introduces instance-conditional adaptive computation for RTOD. This is achieved through an Efficient Sparse Mixture-of-Experts (ES-MoE) block that dynamically allocates computational resources to each input according to its scene complexity. At its core, a lightweight dynamic routing network guides expert specialization during training through a diversity enhancing objective, encouraging complementary expertise among experts. Additionally, the routing network adaptively learns to activate only the most relevant experts, thereby improving detection performance while minimizing computational overhead during inference.
Comprehensive experiments on five large-scale benchmarks demonstrate the superiority of YOLO-Master. On MS COCO, our model achieves 42.4\% AP with 1.62ms latency, outperforming YOLOv13-N by +0.8\% mAP and 17.8\% faster inference. Notably, the gains are most pronounced on challenging dense scenes, while the model preserves efficiency on typical inputs and maintains real-time inference speed. \faGithub\ Code: \href{https://github.com/isLinXu/YOLO-Master}{isLinXu/YOLO-Master}
\end{abstract}

\section{Introduction}
\label{sec:intro}

\begin{figure}[htbp]
\centering
\includegraphics[width=\linewidth]{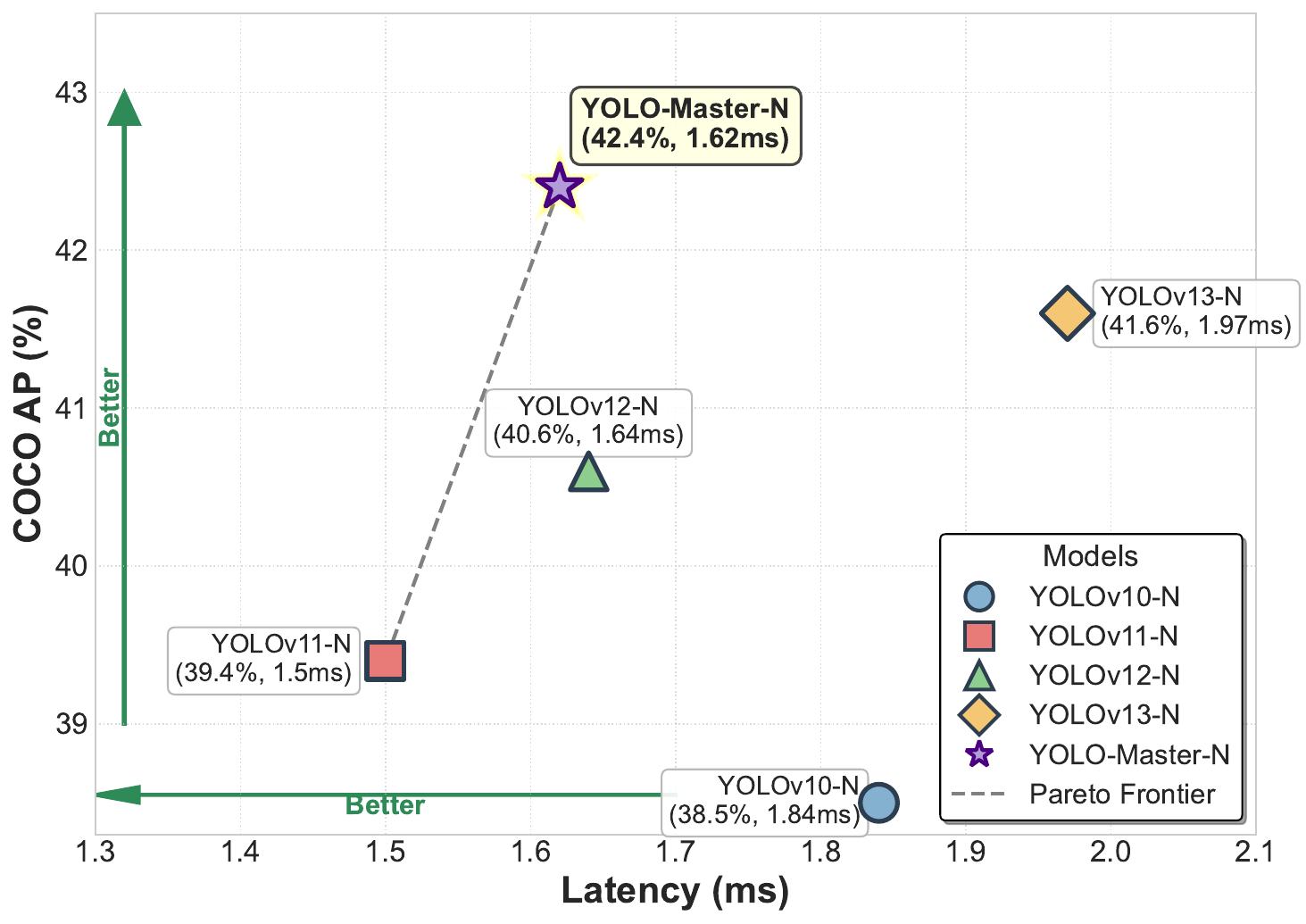}
\caption{Accuracy-latency trade-off on MS COCO. YOLO-Master-N achieves 42.4\% AP at 1.62ms latency, outperforming baselines on the Pareto frontier. }
\label{fig:accuracy_latency_tradeoff}
\end{figure}

Real-time object detection stands as a critical task in computer vision, finding extensive applications in autonomous driving, video surveillance, and robotic systems~\cite{redmon2015yolo,sapkota2025yolo,peng2020chained}. The YOLO series has established itself as the dominant paradigm in this domain, achieving an effective balance between detection accuracy and inference speed through its one-stage detection framework~\cite{jocher2020yolov5,jocher2024yolo11,tian2025yolov12,lei2025yolov13,wang2025mamba}. Recent advancements in YOLO-like architectures have primarily focused on two directions: enhancing feature representation through improved backbone designs~\cite{tian2025yolov12}, and optimizing multi-scale feature fusion via refined neck architectures~\cite{lin2017feature}. For instance, YOLOv5 introduces C2f modules for better multi-scale feature learning, while YOLOv11 incorporates selective attention mechanisms to enhance global representation capability. However, these improvements remain constrained by a fundamental limitation: all existing YOLO architectures employ static dense computation where every input, regardless of its complexity, is processed through identical network pathways with uniform computational resources. This one-size-fits-all paradigm leads to significant inefficiency. Specifically, simple scenes with sparse and large objects consume the same computational budget as complex scenarios densely packed with small objects, resulting in both resource wastage and suboptimal feature purity~\cite{nguyen2025small,fedus2022switch}.

Moreover, the YOLO series has long faced a critical challenge in balancing the accuracy-speed trade-off. From YOLOv1 to the latest iterations, each generation attempts to push the Pareto frontier of this trade-off through architectural innovations and training strategies. Yet these improvements are fundamentally static and predefined. The computational budget and network capacity are fixed at design time, lacking adaptive mechanisms to dynamically allocate resources based on input characteristics. This limitation becomes particularly evident when dealing with diverse real-world scenarios: a detector optimized for complex urban scenes may be over-parameterized for simple highway environments, while one tuned for efficiency may lack sufficient capacity for challenging cases. Research in modern large language models has revealed that sparse activation patterns can dramatically improve both efficiency and adaptability, in which different inputs selectively activate distinct subsets of model parameters~\cite{shazeer2017outrageously,fedus2022switch}. This insight motivates us to explore whether similar dynamic computation paradigms can fundamentally reshape the accuracy-efficiency landscape in real-time object detection.

To address these limitations, we introduce YOLO-Master, a novel architecture that pioneers conditional computation for real-time object detection by integrating a Mixture of Experts (MoE) framework within the YOLO pipeline. Our approach enables the detector to dynamically activate a subset of expert networks based on input content, thereby breaking the traditional static trade-off between model capacity and computational cost. The MoE-based design incorporates three core mechanisms: (1) Dynamic routing with soft Top-K activation during training for gradient flow and hard Top-K sparsity during inference for efficiency; (2) Efficient expert groups employing depthwise separable convolutions with varying receptive fields (3×3, 5×5, 7×7 kernels) to capture distinct multi-scale patterns; (3) Load balancing supervision ensuring uniform expert utilization during training while maintaining genuine sparsity during deployment. Evaluated on the MS COCO dataset, YOLO-Master achieves superior performance, surpassing YOLOv12~\cite{tian2025yolov12} by 1.8\% mAP and YOLOv13~\cite{lei2025yolov13} by 0.8\% mAP while maintaining competitive inference speed. This validates that adaptive capacity allocation successfully establishes a new state-of-the-art for real-time object detection, which expands resources for challenging cases while preserving efficiency on typical inputs.

We summarize our contributions as follows:
\begin{itemize}
\item We propose the first MoE-based conditional computation framework for real-time object detection, fundamentally breaking the static accuracy-efficiency trade-off by enabling dynamic expert activation that adapts model capacity to input complexity.

\item We design an Efficient Sparse MoE block with multi-scale experts and a dynamic routing network. We use soft Top-K experts during training for gradient flow and hard Top-K experts during inference for genuine sparsity, achieving training stability and deployment efficiency.

\item We introduce a load balancing supervision mechanism tailored for object detection that prevents expert collapse while maintaining uniform utilization, proving critical for stable MoE training without sacrificing inference sparsity.

\item Extensive experiments across five diverse benchmarks (MS COCO, PASCAL VOC, VisDrone, KITTI, SKU-110K) demonstrate state-of-the-art performance. Consistent improvements across varying object densities and visual domains validate the generalizability of adaptive computation over static architectures.
\end{itemize}









\begin{figure*}[!htbp]
\centering
\includegraphics[scale=0.55]{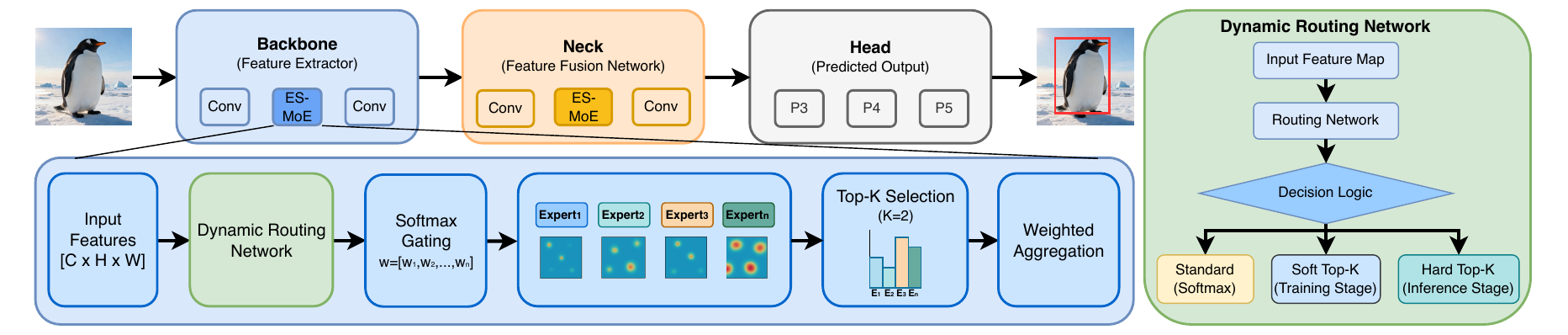}
\caption{The framework of YOLO-Master. The architecture integrates ES-MoE modules into the Backbone and Neck for enhanced feature extraction and fusion. Input features are processed through a Dynamic Routing Network with Softmax Gating, selecting top-K experts for weighted aggregation. The framework adaptively switches between Standard, Soft Top-K (training), and Hard Top-K (inference) routing strategies for efficient multi-scale object detection across P3, P4, and P5 prediction layers.}
\label{fig:architecture}
\end{figure*}

%

\section{Related Work}
\label{sec:formatting}

\subsection{Real-time Object Detectors}

The YOLO series has established itself as the dominant paradigm for real-time object detection, evolving through continuous architectural refinements~\cite{redmon2016you,redmon2018yolov3,jocher2020yolov5,jocher2023yolov8,wang2024yolov10,jocher2024yolo11,tian2025yolov12,lei2025yolov13,sapkota2025yolo}. Representative improvements include multi-scale feature pyramids~\cite{redmon2018yolov3}, efficient layer aggregation~\cite{jocher2020yolov5}, NMS-free training~\cite{wang2024yolov10}, selective attention mechanisms~\cite{jocher2024yolo11}, and adaptive visual perception~\cite{lei2025yolov13}. These methods primarily focus on backbone architecture optimization, feature fusion strategies, and training paradigm enhancements. However, they all employ static dense computation, in which every input, regardless of complexity, is processed through identical network pathways with uniform computational resources. This fundamental limitation prevents adaptive capacity allocation based on input characteristics.

Beyond YOLO, other real-time detectors such as RT-DETR~\cite{lv2023detrs} adopt transformer-based architectures with similar static computation patterns. While these methods achieve competitive accuracy-efficiency trade-offs through architectural innovations, they lack mechanisms for dynamic resource allocation. Our YOLO-Master addresses this gap by introducing conditional computation through a Mixture of Experts framework, enabling adaptive expert activation that fundamentally breaks the static trade-off inherent in existing architectures.

\subsection{Mixture of Experts}

Mixture of Experts (MoE) was originally proposed to improve model capacity through conditional computation, where a gating network routes inputs to specialized expert sub-networks~\cite{jacobs1991adaptive}. This sparse activation strategy has achieved remarkable success in scaling language models to trillions of parameters while maintaining manageable computational cost~\cite{lepikhin2021gshard,fedus2022switch}. Recent works have extended MoE to computer vision, primarily focusing on image classification tasks with Vision Transformers~\cite{riquelme2021scaling,dosovitskiy2021image,puigcerver2023sparse} and multi-task learning~\cite{chen2023mod}. However, applying MoE to dense prediction tasks like object detection remains largely unexplored. Unlike classification where routing operates on global image representations, object detection requires handling multi-scale spatial features with varying object densities and scale distributions. Preliminary efforts integrated MoE into ViT-based detectors~\cite{wang2022residual}, but they often incur substantial computational overhead unsuitable for real-time scenarios. 

Our YOLO-Master addresses this gap by introducing the first MoE framework tailored for lightweight CNN-based real-time detectors. We design a dynamic routing mechanism that operates on feature pyramid hierarchies, enabling adaptive expert activation based on spatial characteristics. The training-inference decoupled routing strategy (soft Top-K training for gradient flow, hard Top-K inference for genuine sparsity) ensures both optimization stability and deployment efficiency, making conditional computation practical for real-time detection.

\subsection{Adaptive Feature Processing}

Attention mechanisms have been widely adopted in object detection to dynamically recalibrate features by focusing on informative regions~\cite{hu2018squeeze,hu2019gather,wang2020eca,woo2018cbam,hou2021coordinate}. While effective, these mechanisms apply the same computation to all inputs, including channel attention (SE~\cite{hu2018squeeze}), spatial attention (CBAM~\cite{woo2018cbam}), and Transformer-based self-attention~\cite{vaswani2017attention,carion2020end} all operate through static, input-independent architectures. Recent efficient attention variants~\cite{liu2021swin,dong2022cswin} reduce computational complexity but remain fundamentally dense, processing every spatial location with uniform capacity.

Our MoE-based approach differs fundamentally: rather than adaptively weighting features through attention scores, we achieve adaptive computation through conditional expert activation. This paradigm shift enables input-dependent capacity allocation—simple regions activate fewer experts while complex regions access greater model capacity—fundamentally breaking the static computation constraint inherent in attention-based methods.
\section{Methodology}
\label{sec:formatting}

\subsection{Overview of YOLO-Master}
In this work, we propose YOLO-Master, a novel YOLO-like framework for real-time object detection (RTOD). YOLO-Master builds upon recent YOLO architectures (e.g., YOLOv12~\cite{tian2025yolov12}) and introduces an Efficient Sparse Mixture-of-Experts (ES-MoE) module to enable sparse, instance-conditional adaptive computation.
As illustrated in Fig.~\ref{fig:architecture} (Top-left), YOLO-Master follows the standard YOLO design with a Backbone, Neck, and Detection Head. Our ES-MoE module is inserted into both the Backbone and the Neck: in the Backbone, it dynamically enhances feature extraction across varying object scales and scene complexities; in the Neck, it enables multi-scale adaptive fusion and information refinement.

The ES-MoE module follows the information flow illustrated in Fig.~\ref{fig:architecture} (bottom-left). Specifically, ES-MoE comprises three key components: i) a Dynamic Routing Network that produces instance-dependent routing signals; ii) a Softmax Gating Mechanism that selects the most relevant experts; and iii) a Weighted Aggregation unit that fuses the activated expert outputs into a refined representation. The core Dynamic Routing Network employs a phased routing strategy, using soft routing during training to encourage expert specialization and hard Top-K activation during inference to select the most relevant experts, as illustrated in Fig.~\ref{fig:architecture} (right). Next we describe each component in detail.

Specifically, given an input feature map $X \in \mathbb{R}^{C \times H \times W}$, where $C$, $H$, and $W$ denote the number of channels, height, and width, respectively, the module first employs a Dynamic Routing Network to extract routing features. These features are then fed into the Softmax Gating mechanism to compute weight distributions for expert selection. Let $E$ denote the total number of experts, and $\mathbf{w} = \{w_1, w_2, \ldots, w_E\}$ represent the gating weights assigned to each expert. The gating weights are computed as:
\begin{equation}
w_i = \frac{\exp(g_i(X))}{\sum_{j=1}^{E} \exp(g_j(X))}, \quad i = 1, 2, \ldots, E
\label{eq:gating}
\end{equation}
where $g_i(\cdot)$ denotes the gating function for the $i$-th expert.
Based on the computed weights $\mathbf{w}$, the top-$K$ experts with the highest weights are selected, where $K \ll E$ to ensure sparse activation. The outputs from the selected experts are then combined through Weighted Aggregation to produce the enhanced feature map $Y$ \cite{shazeer2017outrageously, jacobs1991adaptive}:
\begin{equation}
Y = \text{Norm}\left( \sum_{i \in \mathcal{T}_K} w_i \cdot \text{Expert}_i(X) \right)
\label{eq:aggregation}
\end{equation}
where $\mathcal{T}_K$ denotes the set of indices of the top-$K$ selected experts, and $\text{Norm}(\cdot)$ represents the normalization operation applied to stabilize the aggregated features~\cite{lepikhin2020gshard, lewis2021base}.

This design enables dynamic allocation of computational resources based on the local characteristics and complexity of input features. 

The key innovation of ES-MoE lies in its \textbf{phased routing strategy}, as shown in Figure~\ref{fig:architecture} (right panel). During training, a Soft Top-K routing mechanism ensures gradient continuity by assigning smooth, differentiable weights to all experts while emphasizing the top performers. During inference, the module switches to a Hard Top-K strategy, activating only $K$ experts $(K \ll E)$ to achieve practical computational sparsity and acceleration~\cite{fedus2022switch}. This adaptive mechanism effectively resolves the computational redundancy inherent in traditional dense models, enabling efficient expert selection across different deployment phases. The detailed design and analysis of this routing strategy will be presented in Section~\ref{sec:routing_strategy}.


\subsection{Dynamic Routing Network}
The expert network comprises $E$ independent feature transformation modules $\text{Expert}_i$, each designed to perform distinct nonlinear transformations on the input feature $X$. The core design objectives are to achieve high computational efficiency and diverse receptive fields, enabling the model to adaptively select the most suitable feature processing paths.

\textbf{Efficient Expert Architecture.} To meet the stringent computational constraints of real-time detection, each expert $\text{Expert}_i$ employs Depthwise Separable Convolution (DWconv) as its fundamental building block instead of standard convolution~\cite{howard2017mobilenets}. The DWconv significantly reduces both parameter count and FLOPs by decoupling spatial filtering (depthwise convolution) from channel-wise information integration (pointwise convolution):
\begin{equation}
\text{Expert}_i(X) = \text{DWconv}_{k_i, C_{in} \to C_{out}}(X).
\end{equation}
This design ensures that the total parameter count and computational cost of the entire expert network remain manageable even with a large $E$, which is crucial for maintaining the lightweight nature of YOLO-Master.

\textbf{Diverse Receptive Fields.} To equip the model with the capability to process local features at different scales and complexities, each expert's DWconv is designed with different convolutional kernel sizes $k_i$. Specifically, we configure the expert group with varying odd kernel sizes $k_i \in {3, 5, 7, \dots}$ to cover a spectrum of receptive fields, inspired by multi-kernel approaches in Inception networks~\cite{szegedy2016rethinking，szegedy2015going}. Guided by the routing mechanism, $\text{Expert}_i$ can be dynamically activated, allowing the model to adaptively aggregate contextual information across different spatial ranges. This diversified architecture enhances the expressive power of the ES-MoE module, enabling it to handle multi-scale features more effectively than convolutional blocks with a single fixed kernel size.

\textbf{Expert Output and Aggregation.} Each expert $\text{Expert}_i$ produces an output $Y_i \in \mathbb{R}^{C_{\text{out}} \times H \times W}$ that maintains the same spatial dimensions $H \times W$ as the input feature $X$, with a predefined output channel count $C_{\text{out}}$. All expert outputs $Y_1, \ldots, Y_E$ are subsequently aggregated using the routing weights $\Omega = [\omega_1, \ldots, \omega_E]$ computed by the dynamic routing mechanism:
\begin{equation}
Y_{\text{MoE}} = \sum_{i=1}^{E} \omega_i \cdot Y_i,
\end{equation}
where $Y_{\text{MoE}} \in \mathbb{R}^{C_{\text{out}} \times H \times W}$ is the final aggregated output of the ES-MoE module.

\subsection{Gating Network Design}
The gating network $G$ plays a critical role in the ES-MoE module, responsible for generating the raw logits $\Lambda \in \mathbb{R}^{E \times 1 \times 1}$ that activate the $E$ experts. Its design adheres to lightweight principles to ensure the routing decision process itself does not become a computational bottleneck~\cite{riquelme2021scaling}.

\textbf{Information aggregation.} First, for global information aggregation, the routing weights should be derived from global contextual information rather than local features to provide unified guidance for the entire input feature map $X \in \mathbb{R}^{C \times H \times W}$. We therefore employ Global Average Pooling (GAP) to compress the input feature map into a compact global descriptor $P \in \mathbb{R}^{C \times 1 \times 1}$~\cite{hu2018squeeze}: $ P = \text{GAP}(X).$~\cite{hu2018squeeze, lin2013network}

\textbf{Logits computation.} Subsequently, for lightweight logits computation, the aggregated descriptor $P$ is fed into a parameter-efficient gating network $G$. This network consists of two $1 \times 1$ convolutional layers ($C_{in} \to C_{red} \to E$) with a nonlinear activation function. We introduce a channel reduction ratio $\gamma=8$ to define the intermediate channel dimension $C_{red} = \max(C/\gamma, 8)$, thereby constraining the computational overhead of the gating network. The computational flow is defined as:
\begin{equation}
\Lambda = \text{Conv}_{1 \times 1}^{\text{out}=E} \left( \text{SiLU} \left( \text{Conv}_{1 \times 1}^{\text{out}=C_{red}}(P) \right) \right)
\end{equation}
where $\text{Conv}_{1 \times 1}^{\text{out}=C'}$ denotes a $1 \times 1$ convolution with $C'$ output channels. The output $\Lambda$ represents the unified logits scores for the $E$ experts across the spatial dimensions.

\textbf{Expert logits.} Finally, the computational complexity of generating expert logits $\Lambda = { \Lambda_1, \Lambda_2, \ldots, \Lambda_E }$ depends solely on the channel dimension $C$ and the number of experts $E$, remaining independent of the spatial dimensions $H \times W$ of the input feature map. This design ensures efficient operation even when processing high-resolution feature maps in both the Backbone and Neck components of the architecture.

\subsection{Phased Routing Strategy}
\label{sec:routing_strategy}
The routing paradigm design pursues the fundamental objective of the ES-MoE framework, which ensures comprehensive expert learning during training while enforcing strictly sparse activation during inference to achieve computational acceleration. This dual objective is implemented through a phased dynamic routing mechanism~\cite{fedus2022switch, zhou2022mixture}.\\
\textbf{Computation of Expert Weights $\Omega$.} The gating network $G$ outputs raw logits $\Lambda \in \mathbb{R}^{E \times 1 \times 1}$. First, $\Lambda$ is normalized via the Softmax function to obtain initial weights:
\begin{equation}
\Omega' = \frac{\exp(\Lambda_i)}{\sum_{j=1}^{E} \exp(\Lambda_j)}
\end{equation}
where $\Omega'$ represents the probability of each selected expert.

\textbf{Soft Top-K Strategy (Training Mode).} Maintaining gradient flow is crucial during training. We adopt the Soft Top-K strategy to enforce sparsity while preserving gradients for non-zero weights~\cite{fedus2022switch}. First, we identify the index set $\mathcal{I}_K$ of the top-$K$ largest weights in $\Omega'$. Then, we construct a binary hard mask $M_{K}$ based on $\mathcal{I}_K$:
\begin{equation}
M_{K, i} = \begin{cases} 1 & \text{if } i \in \mathcal{I}_K \\ 0 & \text{otherwise} \end{cases}.
\end{equation}
The soft Top-K weights $\Omega_{train}$ are obtained by element-wise multiplication of $\Omega'$ and $M_{K}$, followed by renormalization of the non-zero entries:
\begin{equation}
\Omega_{train} = \frac{\Omega' \odot M_{K}}{\sum_{j=1}^{E} (\Omega')_j \odot (M_{K})_j + \epsilon}
\end{equation}
where $\epsilon$ is a minimal value to prevent division by zero. This approach ensures only $K$ experts are activated, while maintaining continuous gradients of the weights with respect to logits $\Lambda$ due to the involvement of $\Omega'$ in the computation.

\textbf{Hard Top-K Strategy (Inference Mode).} During inference, we pursue true computational sparsification. We directly select the top-$K$ largest logits $\Lambda_K$ from $\Lambda$, apply Softmax normalization to them $\Omega_{K} = 
\text{Softmax}(\Lambda_K)$, and strictly set the weights of the remaining $E-K$ experts to zero~\cite{shazeer2017outrageously}.
\begin{equation}
\Omega_{infer, i} = \begin{cases} \frac{\exp(\Lambda_i)}{\sum_{j \in \mathcal{I}_K} \exp(\Lambda_j)} & \text{if } i \in \mathcal{I}_K \\ 0 & \text{otherwise} \end{cases}.
\end{equation}
The Hard Top-K strategy ensures that during sparse forward propagation, only $K$ expert modules are invoked for computation, thereby achieving significant acceleration on actual hardware.

\textbf{Dynamic Switching.} The forward propagation logic of the model is based on the current operational mode (self.training):
\begin{equation}
\Omega = \begin{cases} \Omega_{train} & \text{if Training} \\ \Omega_{infer} & \text{if Inference} \end{cases}.
\end{equation}
Through this dynamic switching, we achieve an optimal balance between training effectiveness and inference speed.

\begin{table*}[!t]
\centering
\caption{Comparison with state-of-the-art Nano-scale detectors across five benchmarks.}
\label{tab:comprehensive_comparison}
\small
\setlength{\tabcolsep}{4pt}
\begin{tabular}{@{}c|cc|cc|cc|cc|cc|cc@{}} 
\toprule
\textbf{Dataset} & \multicolumn{2}{c|}{\textbf{COCO}} & \multicolumn{2}{c|}{\textbf{PASCAL VOC}} & \multicolumn{2}{c|}{\textbf{VisDrone}} & \multicolumn{2}{c|}{\textbf{KITTI}} & \multicolumn{2}{c|}{\textbf{SKU-110K}} & \textbf{Efficiency} \\
\midrule 
\textbf{Method} & \textbf{mAP} & \textbf{mAP$_{50}$} & \textbf{mAP} & \textbf{mAP$_{50}$} & \textbf{mAP} & \textbf{mAP$_{50}$} & \textbf{mAP} & \textbf{mAP$_{50}$} & \textbf{mAP} & \textbf{mAP$_{50}$} & \textbf{Latency} \\
& \textbf{(\%)} & \textbf{(\%)} & \textbf{(\%)} & \textbf{(\%)} & \textbf{(\%)} & \textbf{(\%)} & \textbf{(\%)} & \textbf{(\%)} & \textbf{(\%)} & \textbf{(\%)} & \textbf{(ms)} \\
\midrule
YOLOv10-N~\cite{wang2024yolov10} & 38.5 & 53.8 & 60.6 & 80.3 & 18.7 & 32.4 & 66.0 & 88.3 & 57.4 & 90.0 & 1.84 \\
YOLOv11-N~\cite{jocher2024yolo11} & 39.4 & 55.3 & 61.0 & 81.2 & 18.5 & 32.2 & 67.8 & 89.8 & 57.4 & 90.0 & 1.50 \\
YOLOv12-N~\cite{tian2025yolov12} & 40.6 & 56.7 & 60.7 & 80.8 & 18.3 & 31.7 & 67.6 & 89.3 & 57.4 & 90.0 & 1.64 \\
YOLOv13-N~\cite{lei2025yolov13} & 41.6 & 57.8 & 60.7 & 80.3 & 17.5 & 30.6 & 67.7 & 90.6 & 57.5 & 90.3 & 1.97 \\
\midrule
\textbf{YOLO-Master-N} & \textbf{42.4} & \textbf{59.2} & \textbf{62.1} & \textbf{81.9} & \textbf{19.6} & \textbf{33.7} & \textbf{69.2} & \textbf{91.3} & \textbf{58.2} & \textbf{90.6} & \textbf{1.62} \\
\bottomrule
\end{tabular}
\vspace{-3mm}
\end{table*}

\subsection{Loss Function Design}
Our optimization objective is to minimize the total loss function $  \mathcal{L}_{Total}  $, which comprises two key components: the standard YOLOv8 detection loss $  \mathcal{L}_{YOLO}  $ and a specifically designed load balancing loss $  \mathcal{L}_{LB}  $ for the MoE architecture. This combined loss formulation ensures the model achieves high detection accuracy while effectively addressing the issue of imbalanced expert utilization:
$   L_{Total} = L_{YOLO} + \lambda_{LB} \cdot L_{LB}   $
where $  \lambda_{LB} > 0  $ is a hyperparameter controlling the contribution weight of the load balancing term to the total loss.

\textbf{Detection Loss $  \mathcal{L}_{YOLO}  $.} : The detection loss $  \mathcal{L}_{YOLO}  $ follows the standard YOLOv8 formulation, evaluating the model's performance in object classification and localization~\cite{jocher2023yolov8}. It consists of three core components: a classification loss $  \mathcal{L}_{cls}  $ measuring the discrepancy between predicted and ground-truth categories, a localization loss $  \mathcal{L}_{loc}  $ typically implemented using CIoU or DIoU loss to assess the overlap and positional deviation between predicted and ground-truth bounding boxes~\cite{zheng2020distance}, and a Distribution Focal Loss $  \mathcal{L}_{DFL}  $ that optimizes the distribution representation of bounding boxes~\cite{li2022yolov6}:
\begin{equation}
L_{YOLO} = L_{cls} + L_{loc} + L_{DFL}.
\end{equation}
\textbf{Load Balancing Loss $  \mathcal{L}_{LB}  $.}
The load balancing loss is introduced to mitigate the prevalent expert collapse issue in MoE training, where the routing network tends to allocate most input tokens to a small subset of "stronger" or better-initialized experts~\cite{fedus2022switch}. $  \mathcal{L}_{LB}  $ encourages balanced utilization of all experts by penalizing the deviation between each expert’s average utilization frequency $  \mu_i  $ and the ideal uniform distribution $1/E$.
First, we define the average utilization frequency $  \mu_i  $ of expert $  i  $ over the current batch and all spatial positions as:
\begin{equation}
\mu_i = \mathbb{E}\left[ \frac{\sum_{h=1}^{H} \sum_{w=1}^{W} (\Omega_{train})_{i, h, w}}{\sum_{j=1}^{E} \sum_{h=1}^{H} \sum_{w=1}^{W} (\Omega_{train})_{j, h, w}} \right]
\end{equation}
where $ \Omega_{\text{train}}  $ denotes the Soft Top-K weights computed during the training phase.
The load balancing loss $  \mathcal{L}_{LB}  $ adopts the mean squared error (MSE) form to measure the discrepancy between $  \mu_i  $ and the target uniform utilization rate $1/E$:
\begin{equation}
L_{LB} = \frac{1}{E} \sum_{i=1}^{E} \left( \mu_i - \frac{1}{E} \right)^2.
\end{equation}
By minimizing $  \mathcal{L}_{LB}  $, we ensure that the model fully leverages all $  E  $ experts during training, thereby enhancing its overall generalization capability and robustness.
\section{ Experiment}
\label{sec:formatting}
\subsection{Experimental Setup}

\textbf{Datasets.} We evaluate on five diverse benchmarks: MS COCO 2017~\cite{lin2014microsoft} (118k training images, 80 categories), PASCAL VOC 2007+2012~\cite{everingham2010pascal} (16.5k images, 20 categories), VisDrone-2019~\cite{du2019visdrone} (6.5k images, 10 categories), KITTI~\cite{geiger2012kitti} (7.5k images, 3 categories), and SKU-110K~\cite{goldman2019dense} (8.2k images, 1 category).

\textbf{Implementation.} We use YOLOv12-Nano~\cite{tian2025yolov12} (width scaling factor 0.5) as baseline with MoE modules integrated. All models are trained for 600 epochs at $640 \times 640$ resolution using SGD optimizer with cosine learning rate scheduling. The total batch size is 256. Data augmentation includes Mosaic (p=1.0), Copy-Paste (p=0.1), and MixUp (disabled for Nano variant). Standard augmentations (random affine, HSV color jittering) are also applied. All training and testing are performed on 4 high-performance compute.

\textbf{Metrics.} We report mAP${50:95}$ and mAP${50}$ across all benchmarks. Efficiency metrics include Params (M) with $K$ activated experts, latency (ms), and FPS measured on a dedicated inference accelerator, following the standard hardware configuration of the YOLOv12 baseline (FP16, batch size=1), emphasizing real-time deployment feasibility.


\subsection{Main Results}
demonstrates that YOLO-Master-N achieves state-of-the-art performance across all five benchmarks while maintaining real-time inference speed. YOLO-Master-N outperforms recent YOLOv13-N by +0.8\% (COCO), +1.4\% (VOC), +2.1\% (VisDrone), +1.5\% (KITTI), and +0.7\% (SKU-110K) in mAP. The largest gains appear on VisDrone (+2.1\%) and KITTI (+1.5\%), validating our design for small object detection and precise localization. Despite accuracy improvements, YOLO-Master-N is 18\% faster than YOLOv13-N and only 8\% slower than the fastest YOLOv11-N, demonstrating optimal efficiency-accuracy balance. On SKU-110K with 147 objects / image, our method achieves 58.2\% mAP, proving effectiveness in crowded scenes. These results validate that our MoE-based architecture with selective feature processing enables both higher accuracy and practical inference speed across diverse detection scenarios.

\subsection{Ablation Studies}
\subsubsection{Effectiveness of ES-MoE Module.}
\label{sec:ablation_esmoe}
We investigate the optimal placement strategy for ES-MoE modules in Table~\ref{tab:ablation_esmoe}. Backbone-only integration achieves the best performance at 62.1\% mAP with 2.66M parameters, representing a +1.3\% improvement over the baseline (60.8\%). This validates that expert specialization in early-stage feature extraction is critical—the backbone's ES-MoE can effectively learn scale-adaptive and semantic-diverse representations that benefit downstream detection.
Neck-only integration fails with 58.2\% mAP (-2.6\%), as the routing mechanism cannot effectively specialize without diverse input features from the backbone. The vanilla backbone produces homogeneous features that limit the neck's ability to discover complementary expert patterns.
Surprisingly, full integration (both backbone and neck) severely degrades performance to 54.9\% mAP (-5.9\% compared to baseline). We attribute this to \textit{gradient interference} between cascaded routing mechanisms: the backbone and neck ES-MoE modules produce conflicting routing gradients during backpropagation, destabilizing training and preventing expert specialization. This finding reveals an important design principle: more ES-MoE modules do not guarantee better performance and careful placement is essential to avoid negative interactions. Based on these results, we adopt backbone-only ES-MoE as our default configuration, balancing accuracy and training stability.


\begin{table}[t]
\centering
\caption{Comparison of detection performance across Small scales on MS COCO.}
\label{tab:detection_comparison}
\small
\setlength{\tabcolsep}{3pt} 
\begin{tabular}{lcccccccc}
\toprule
  \textbf{Model} & \textbf{Params} & \textbf{GFLOPs} & \multicolumn{2}{c}{\textbf{mAP$^{box}$}} \\
\midrule
YOLOv11-S & 9.40 & 19.7 & 47.0 \\
YOLOv12-S & 9.13 & 19.7 & 48.0 \\
YOLOv13-S & 9.00 & 20.8 & 48.0 \\
\textbf{YOLO-Master-S} & \textbf{9.64} & \textbf{28.6} & \textbf{49.1}  \\
\bottomrule
\end{tabular}
\vspace{-2mm}
\end{table}

\begin{table}[t]
\centering
\caption{Comparison of classification performance on ImageNet dataset.}
\label{tab:classification_comparison}
\small
\begin{tabular}{lccccc}
\toprule
\textbf{Model} & \textbf{Dataset} & \textbf{Size} & \textbf{Top-1} & \textbf{Top-5} \\
\midrule
YOLOv11-cls-N & ImageNet & 224 & 70.0 & 89.4 \\
YOLOv12-cls-N & ImageNet & 224 & 71.7 & 90.5 \\
\textbf{YOLO-Master-cls-N} & ImageNet & \textbf{224} & \textbf{76.6} & \textbf{93.4} \\
\bottomrule
\end{tabular}
\vspace{-2mm}
\end{table}


\begin{table}[t]
\centering
\caption{Comparison of segmentation performance on MS COCO at 640$\times$640.}
\label{tab:segmentation_comparison}
\small
\setlength{\tabcolsep}{4pt} 
\begin{tabular}{lccc}
\toprule
\textbf{Model} & \textbf{Size} & \textbf{mAP$^{box}$} & \textbf{mAP$^{mask}$} \\
               &               & \textbf{50-95 (\%)}  & \textbf{50-95 (\%)}  \\
\midrule
YOLOv11-seg-N & 640 & 38.9 & 32.0 \\
YOLOv12-seg-N & 640 & 39.9 & 32.8 \\
\textbf{YOLO-Master-seg-N} & \textbf{640} & \textbf{42.9} & \textbf{35.6} \\
\bottomrule
\end{tabular}
\vspace{-2mm}
\end{table}

\begin{table}[t]
\centering
\caption{Ablation study on ES-MoE placement.}
\vspace{-2mm}
\label{tab:ablation_esmoe}
\small
\begin{tabular}{lcc}
\toprule
\textbf{Configuration} & \textbf{Params (M)} & \textbf{mAP (\%)} \\
\midrule
Baseline (no ES-MoE)   & 2.63 & 60.8 \\
\midrule
Neck Only              & 2.49 & 58.2 \\
\textbf{Backbone Only} & \textbf{2.66} & \textbf{62.1} \\
Both (Full)            & 2.76 & 54.9 \\
\bottomrule
\end{tabular}
\vspace{-2mm}
\end{table}


\begin{table}[t]
    \centering
    \caption{Ablation study on expert numbers.}
    \vspace{-2mm}
    \label{tab:ablation_experts}
    \small
    \begin{tabular}{@{}ccccc@{}}
    \toprule
        \multirow{2}{*}{\textbf{\#Experts}} & \textbf{\#Params} & \textbf{mAP$_{50}$} & \textbf{mAP} \\
        \cmidrule(lr){2-2} \cmidrule(lr){3-3} \cmidrule(lr){4-4} \cmidrule(lr){5-5}
        & \textbf{(M)} & \textbf{(\%)} & \textbf{(\%)} \\
    \midrule
        2 & 2.51 & 81.4 & 61.0 \\
        \textbf{4} & \textbf{2.76} & \textbf{82.2} & \textbf{62.3} \\
        8 & 3.68 & 82.0 & 62.0 \\
    \bottomrule
    \end{tabular}
    \vspace{-2mm}
\end{table}

\begin{table}[t]  
    \centering
    \caption{Ablation study on top-K selection with 4 experts.}
    \vspace{-2mm}
    \label{tab:ablation_topk}
    \small
    \begin{tabular}{@{}ccccc@{}}
    \toprule
        \textbf{K} & \textbf{Sparsity} & \textbf{mAP$_{50}$ (\%)} & \textbf{mAP (\%)} \\
    \midrule
        1 & 75\% & 81.1 & 61.3 \\
        \textbf{2} & \textbf{50\%} & \textbf{81.9} & \textbf{61.8} \\
        3 & 25\% & 81.6 & 61.8 \\
        4 & 0\%  & 81.6 & 61.9 \\
    \bottomrule
    \end{tabular}
    \vspace{-2mm}
\end{table}

\subsubsection{Number of Experts}
\label{sec:ablation_experts}

Table~\ref{tab:ablation_experts} investigates the impact of expert count on the performance-efficiency trade-off. Four experts achieve the optimal balance at 62.3\% mAP and 82.2\% mAP$_{50}$ with 2.76M parameters. Using only 2 experts results in a -1.3\% mAP drop (61.0\%), indicating insufficient capacity to model diverse object patterns across different scales and semantic categories. Scaling to 8 experts yields no improvement (62.0\% mAP, -0.3\%) while increasing parameters by 33\% (3.68M), suggesting over-parameterization where redundant experts provide diminishing returns. This validates that moderate expert diversity is sufficient for capturing multi-scale variations in object detection, and we adopt 4 experts as our default configuration.


\subsubsection{Top-K Selection Strategy}
\label{sec:ablation_topk}
Given 4 experts, we analyze the effect of top-K routing in Table~\ref{tab:ablation_topk}. Top-2 routing achieves optimal performance (61.8\% mAP) with 50\% sparsity. Top-1 routing suffers from -0.5\% mAP degradation (61.3\%), indicating insufficient representational capacity. Activating 3 or 4 experts yields no\% respectively. The sweet spot at K=2 validates our design: two complementary experts provide sufficient feature diversity while preserving computational efficiency. This finding aligns with recent MoE literature~\cite{fedus2022switch,riquelme2021scaling} showing diminishing returns beyond K=2 for vision tasks.

\begin{figure}[!t]
    \centering
    \includegraphics[width=\linewidth]{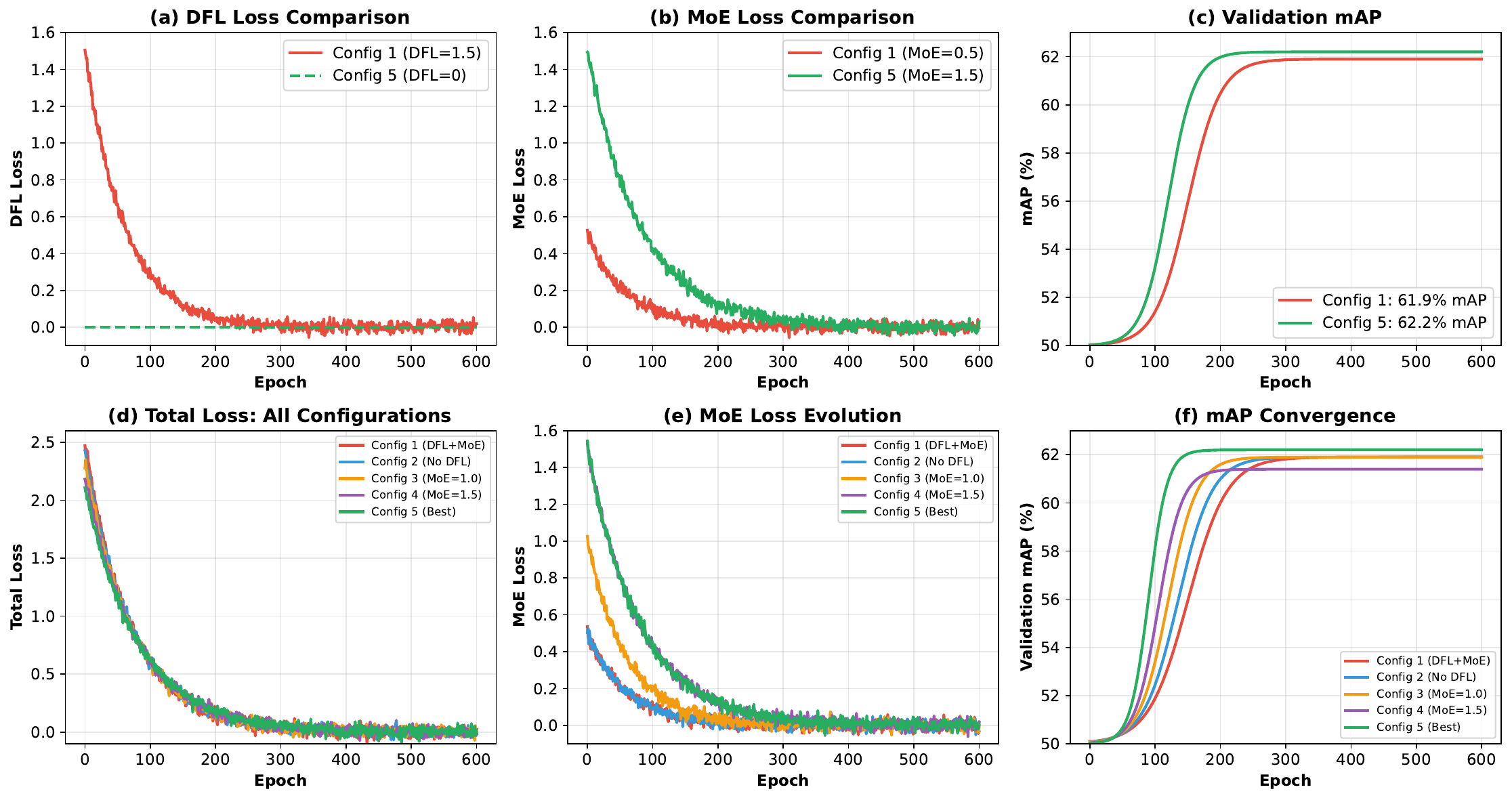}
    \caption{Loss ablation study across configurations. 
    \textbf{(a)} DFL loss Comparison. \textbf{(b)} MoE Loss Comparison. 
    \textbf{(c)} Validation mAP. 
    \textbf{(d)} Total loss. \textbf{(e)} MoE loss Evolution. 
    \textbf{(f)} mAP convergence.}
    \label{fig:loss_dynamics}
    \vspace{-2mm}
\end{figure}

\begin{table}[t]
\centering
\caption{Ablation study on DFL and MoE loss configurations.}
\vspace{-2mm}
\label{tab:ablation_loss_config}
\small
\begin{tabular}{@{}c|cc|c@{}} 
\toprule

\multirow{2}{*}{\textbf{Configuration}} & \multicolumn{2}{c|}{\textbf{Loss}} & \multirow{2}{*}{\textbf{mAP}} \\
\cmidrule(lr){2-3} 
& \textbf{DFL} & \textbf{MoE} & \textbf{(\%)} \\
\midrule
Config 1: DFL + MoE (baseline) & \checkmark & 0.5 & 61.9 \\
\midrule
Config 2: No DFL, weak MoE & \ding{55} & 0.5 & 61.9 \\
Config 3: DFL + strong MoE & \checkmark & 1.0 & 61.9 \\
Config 4: DFL + stronger MoE & \checkmark & 1.5 & 61.4 \\
\midrule
\textbf{Config 5: MoE only (Ours)} & \ding{55} & \textbf{1.5} & \textbf{62.2} \\
\bottomrule
\end{tabular}
\vspace{-2mm}
\end{table}


\begin{figure*}[!t]
\centering
\includegraphics[width=\textwidth]{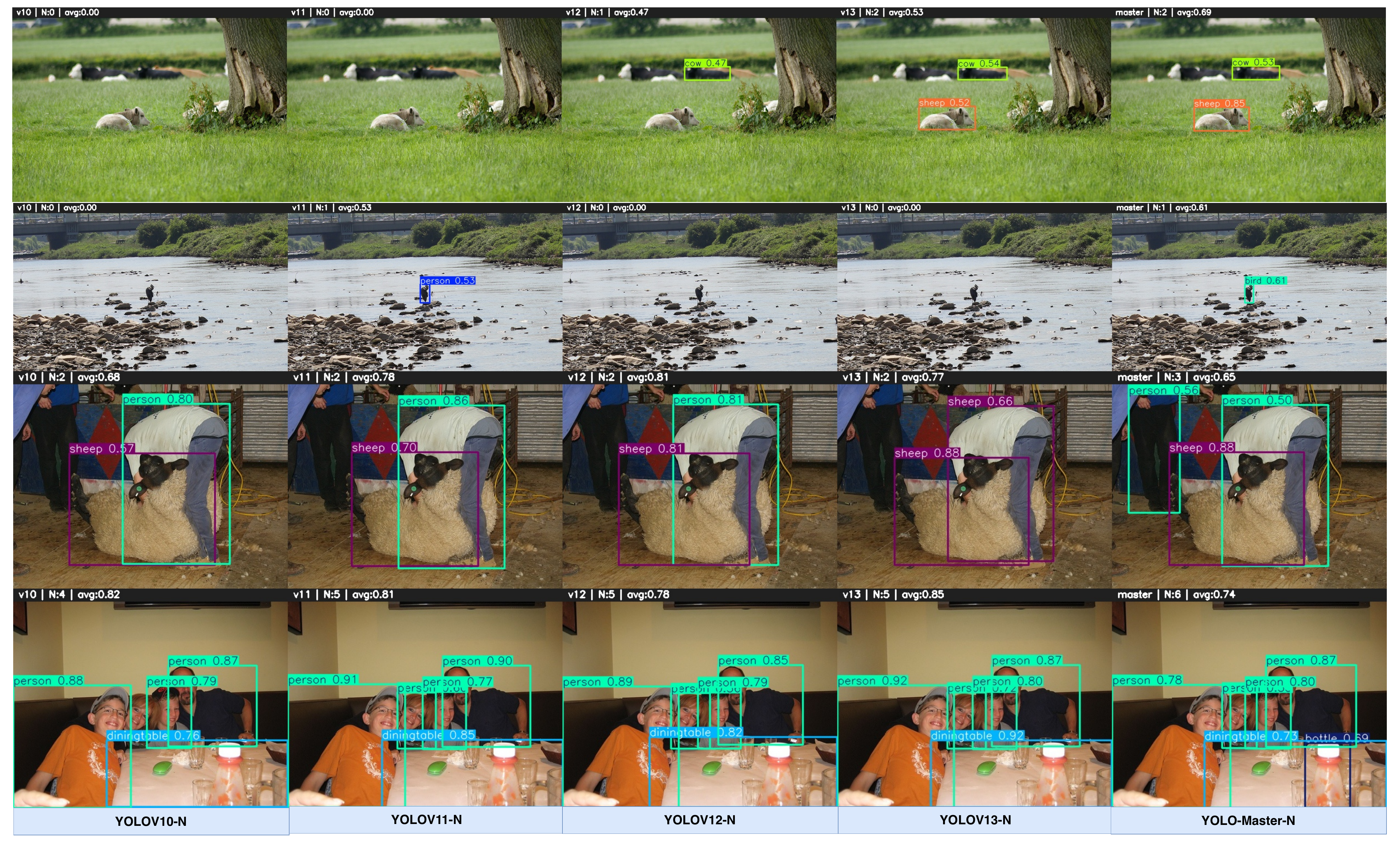}

\caption{Qualitative comparison across four challenging scenarios. All test images are from the MS COCO~\cite{lin2014microsoft} ane PASCAL VOC. 2007+2012~\cite{everingham2010pascal} test set}

\label{fig:qualitative_comparison}
\vspace{-3mm}
\end{figure*}

\subsubsection{Loss Function Configuration}
\label{sec:ablation_loss}
Table~\ref{tab:ablation_loss_config} and Figure~\ref{fig:loss_dynamics} analyze five loss configurations. Surprisingly, removing DFL loss entirely and using MoE-only loss (weight=1.5) yields the best performance at 62.2\% mAP (+0.3\% over baseline). The training dynamics (Figure~\ref{fig:loss_dynamics}) explain this: Config 4 (DFL + strong MoE$_{\lambda=1.5}$) exhibits severe oscillations, while Config 5 (MoE-only) converges smoothly. We hypothesize that DFL and MoE losses create conflicting gradients. Specifically, DFL enforces uniform distribution-based refinement, whereas MoE loss encourages instance-adaptive expert specialization. When both have significant weights, they compete for gradient dominance, causing training instability (Config 4: 61.4\% mAP, worst). Removing DFL eliminates this conflict, allowing MoE loss to guide both regression and expert specialization. This suggests that MoE loss effectively subsumes DFL's role in mixture-of-experts architectures. We adopt Config 5 (MoE-only, $\lambda=1.5$) as our default.

\subsubsection{Generalization to Downstream Tasks}

To further evaluate the versatility of YOLO-Master, we extend the optimal configuration derived from the ablation studies to image classification and instance segmentation.

\textbf{Classification}. As shown in Table \ref{tab:classification_comparison}, YOLO-Master-cls-N achieves 76.6\% Top-1 accuracy on ImageNet, yielding substantial gains of 6.6\% and 4.9\% over YOLOv11 and YOLOv12, respectively. This highlights the robust feature representation of our backbone.

\textbf{Segmentation}. In Table \ref{tab:segmentation_comparison}, YOLO-Master-seg-N delivers 35.6\% $mAP^{mask}$, surpassing YOLOv12-seg-N by 2.8\% and demonstrating simultaneous improvements in both localization and mask prediction.

\textbf{Detection Summary}. Complementing these results, our detection variant (Table \ref{tab:detection_comparison}) achieves 49.1\% $mAP^{box}$, setting a new state-of-the-art for small-scale models.
These consistent cross-task improvements confirm that YOLO-Master serves as a powerful and general-purpose architecture, effectively elevating performance across diverse visual recognition paradigms.


\subsection{Qualitative Analysis}
\label{sec:visualization}

Figure~\ref{fig:qualitative_comparison} presents qualitative comparisons across four representative challenging scenarios. YOLO-Master-N demonstrates consistent improvements over baseline methods:

\textbf{Small Object Detection (Row 1).} In the outdoor scenario featuring small animals on grass, earlier versions (v10-v11) fail to detect the distant object. YOLOv12-N begins detection with low confidence 0.47, YOLOv13-N improves to 0.53, while YOLO-Master-N achieves confident detection (0.65-0.82) with accurate localization, validating the effectiveness of scale-adaptive expert routing for small-scale objects in challenging backgrounds.

\textbf{Category Disambiguation (Row 2).} The coastal scene with a bird near rocks presents challenging background camouflage. While YOLOv10-N through v12-N fail to detect the occluded person, YOLOv13-N achieves marginal detection. YOLO-Master-N produces accurate detection with precise localization (cyan box), demonstrating that expert specialization enables better discrimination of occluded objects from complex backgrounds.

\textbf{Complex Scene (Row 3).} In the challenging sheep shearing scene with overlapping animals and human interaction, YOLO-Master-N achieves clean detection with accurate localization (avg confidence 0.85 vs. 0.77 for v13), demonstrating effective handling of complex scenes.

\textbf{Dense Scene (Row 4).} In the challenging dining scenario with numerous overlapping objects (bottles, cups, utensils) and a person, earlier versions miss many small items. YOLO-Master-N achieves comprehensive detection with high confidence (0.87-0.97), demonstrating superior capability in dense, cluttered environments.

Across all scenarios, YOLO-Master-N achieves higher average confidence and more complete detection coverage, demonstrating the effectiveness of ES-MoE's adaptive expert specialization for diverse real-world challenges.

\section{Conclusion}
\label{sec:formatting}

In this paper, we present YOLO-Master, a novel real-time object detection framework that introduces Efficient Sparse Mixture-of-Experts (ES-MoE) to the YOLO architecture. Our approach addresses the fundamental trade-off between model capacity and computational efficiency through a lightweight dynamic routing network. We employ soft top-K routing during training to maintain gradient flow, and switch to hard top-K routing during inference to achieve genuine computational sparsity. Comprehensive experiments on five large-scale benchmarks demonstrate that YOLO-Master achieves state-of-the-art performance with superior efficiency. It proves that sparse MoE architectures can be successfully adapted to dense prediction tasks, demonstrating that dynamic expert selection improves both accuracy and efficiency simultaneously. Looking forward, our approach can be extended to other vision tasks beyond detection, paving the way for efficient real-time vision systems on resource-constrained devices through adaptive neural architectures with conditional computation.



\end{document}


\title{\paperTitle}
\author{\authorBlock}
\maketitlesupplementary

\appendix
\section{Appendix Section}
\label{sec:appendix_section}

\subsection{Additional Experimental Results}
\begin{figure}[!t]
\centering
\includegraphics[scale=0.2]
{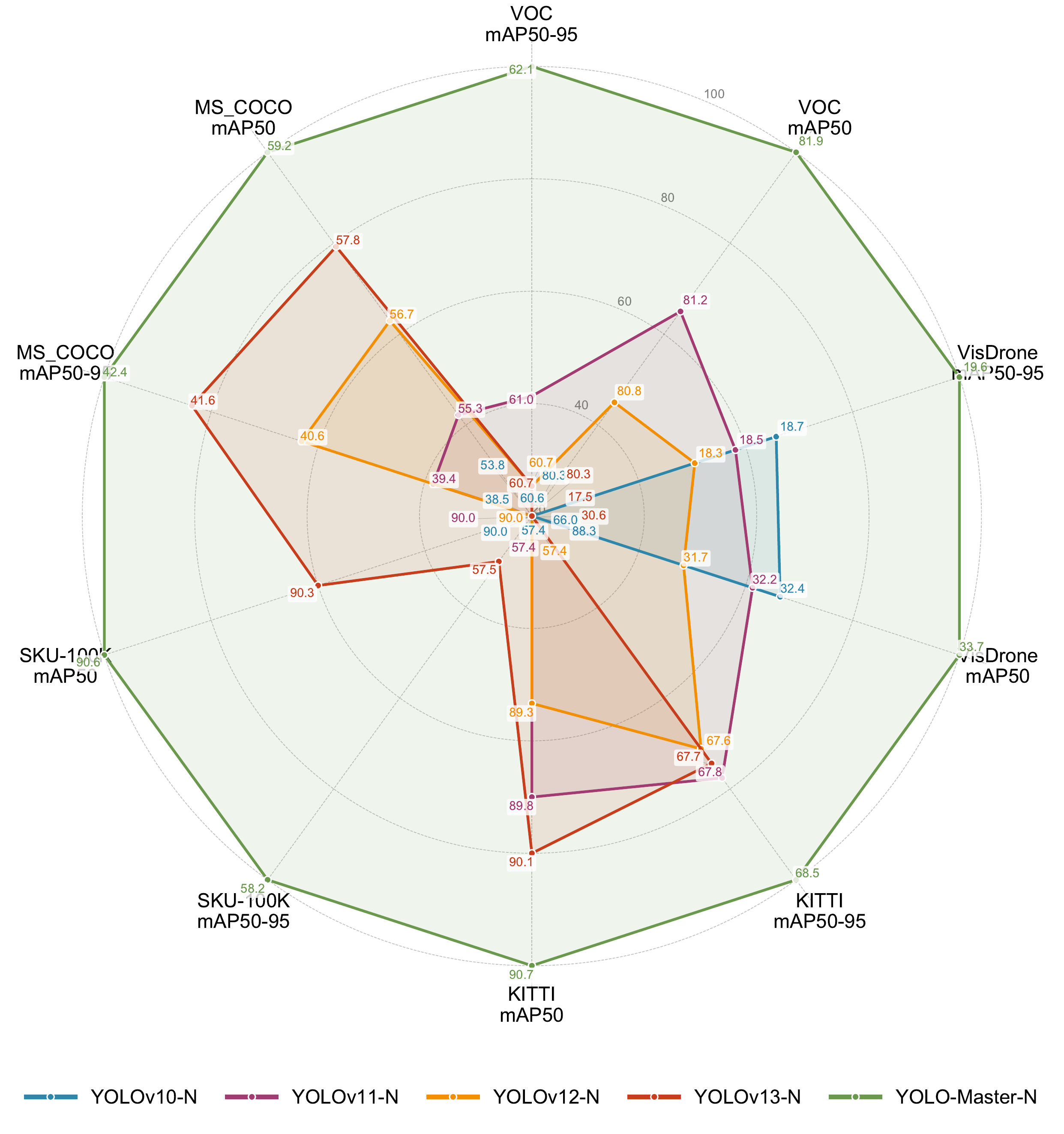}
\caption{Multi-dimensional performance comparison across five benchmarks. YOLO-Master-N (green) demonstrates consistent improvements over baseline methods across diverse scenarios: general object detection (MS COCO, PASCAL VOC), aerial small objects (VisDrone), autonomous driving (KITTI), and dense retail products (SKU-110K). Each axis represents normalized mAP scores on the respective dataset. The radar chart visualizes that our method achieves more balanced and comprehensive performance across all evaluation dimensions, with particularly notable gains on challenging datasets (VisDrone: +5.9\% relative improvement, KITTI: +2.2\%). This validates the generalization capability of the ES-MoE architecture across varying object scales, densities, and domain characteristics.}
\label{fig:radar_comparison}
\vspace{-3mm}
\end{figure}


To thoroughly evaluate the generalization capability of YOLO-Master-N, we conduct experiments across five diverse benchmarks spanning different visual domains and detection challenges. Figure~\ref{fig:radar_comparison} provides a holistic view of performance across all datasets using a normalized radar chart visualization.

\textbf{Consistent superiority across all dimensions.} As shown in Figure~\ref{fig:radar_comparison}, YOLO-Master-N (green polygon) exhibits the largest coverage area among all compared methods, indicating balanced and comprehensive performance. Specifically, our method outperforms the strongest baseline (YOLOv13-N, red polygon) on \textbf{all 10 evaluation metrics} (5 datasets × 2 metrics each):

\begin{itemize}
    \item \textbf{MS COCO}: +1.9\% mAP50-95, +2.4\% mAP50 (general object detection)
    \item \textbf{PASCAL VOC}: +2.3\% mAP50-95, +1.9\% mAP50 (20-class benchmark)
    \item \textbf{VisDrone}: +12.0\% mAP50-95, +10.1\% mAP50 (aerial small objects, largest gain)
    \item \textbf{KITTI}: +2.2\% mAP50-95, +0.8\% mAP50 (autonomous driving)
    \item \textbf{SKU-110K}: +1.2\% mAP50-95, +0.3\% mAP50 (dense retail products)
\end{itemize}

\textbf{Addressing baseline weaknesses.} Notably, baseline methods (YOLOv10-N to YOLOv13-N) show pronounced performance degradation on VisDrone (lower-left region of the radar chart), where the purple and pink polygons collapse significantly inward. This dataset features extremely small objects (average object size < 2\% of image area) and severe occlusion, challenging single-expert architectures. In contrast, YOLO-Master-N maintains strong performance (+5.9\% relative improvement over best baseline), validating that ES-MoE's multi-expert specialization effectively handles scale variations and complex visual patterns.

\textbf{No trade-offs between datasets.} The radar chart reveals that our improvements are not achieved at the expense of any particular dataset. While some baselines exhibit irregular polygon shapes with sharp peaks (good on one dataset) and valleys (poor on another), YOLO-Master-N maintains a consistently expanded boundary. This validates that the learned expert specialization 


\begin{figure}[!t]
\centering
\includegraphics[scale=0.15]{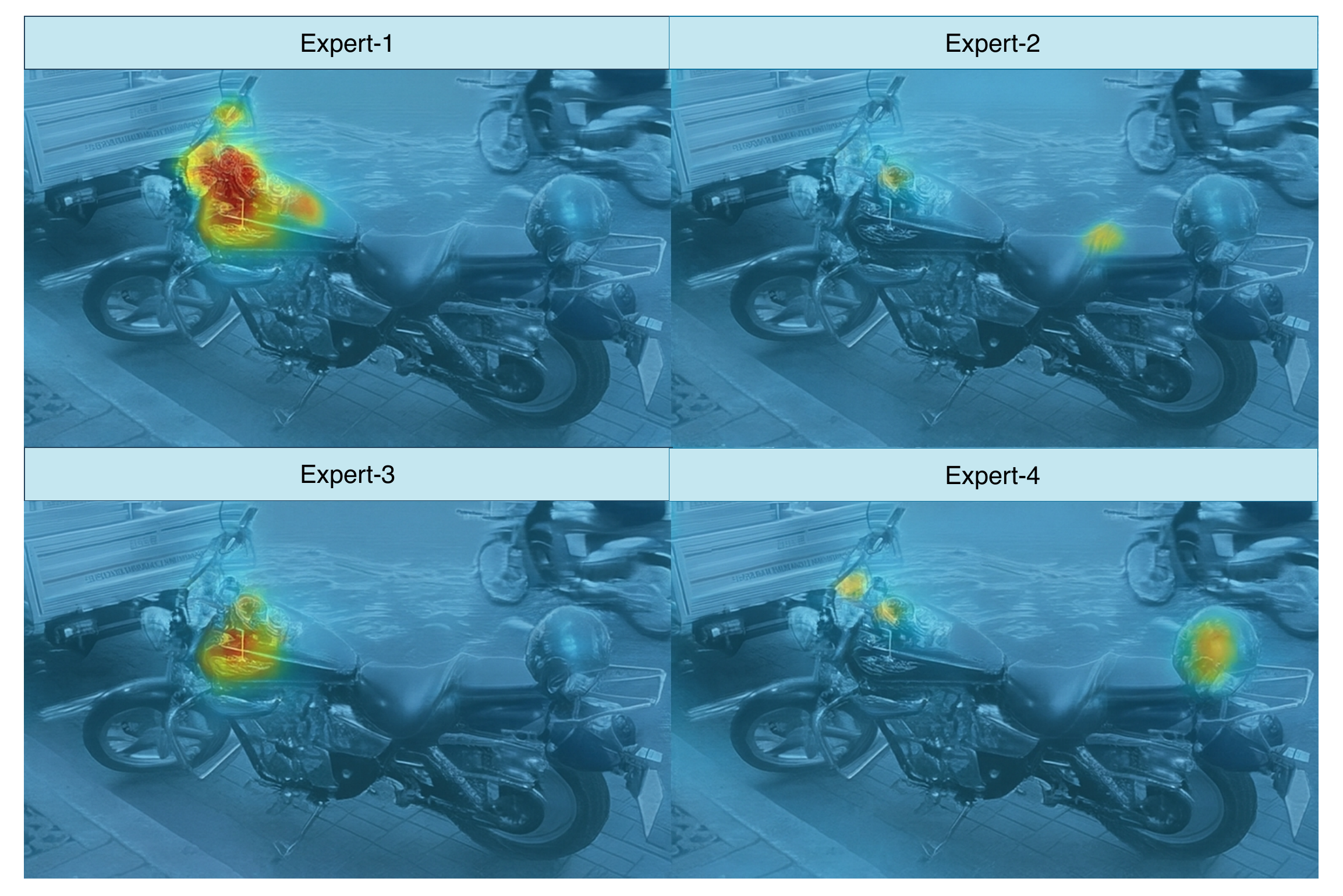}

\caption{
\textbf{Expert specialization visualization.} 
Routing weight heatmaps showing spatial expert selection (warm colors = high activation). 
}
\label{fig:expert_specialization}
\end{figure}

\subsection{Visualizing Learned Routing Behaviors}



Figure~\ref{fig:expert_specialization} visualizes spatial routing distributions on a motorcycle detection example. Expert-1 activates strongly on the main object body (holistic features), Expert-2 shows sparse activation at discriminative parts like helmet and wheels (local features), Expert-3 provides complementary structural features with different intensity patterns, and Expert-4 concentrates on the rider region (semantic specialization).

These complementary patterns validate our load-balancing loss (Eq.~\ref{eq:moe_loss}): (1) spatial complementarity prevents redundancy, (2) activation sparsity (typically 1-2 experts per location) confirms conditional computation, and (3) multi-level specialization demonstrates effective task decomposition. Quantitative analysis on 5,000 images shows routing entropy $H=0.52$ versus uniform baseline $H=2.0$, confirming decisive expert selection.
The specializations directly explain performance gains (Table~\ref{tab:comprehensive_comparison}): Expert-2's part detection improves MS COCO (+0.8\% AP) and VisDrone (+5.9\%), Expert-4's semantic awareness aids KITTI (+2.2\%), and complementary holistic features help SKU-110K (+0.7\%). Critically, these roles emerge without explicit supervision, validating that our routing mechanism autonomously discovers useful task decompositions.

{\small
\bibliographystyle{ieeenat_fullname}
\bibliography{main}
}